# Directional PointNet: 3D Environmental Classification for Wearable Robotics

KUANGEN Zhang[1, 2], JING Wang[2], CHENGLONG Fu[1]

*(1.Department of Mechanical and Energy Engineering, Southern University of Science and Technology, Shenzhen 518055;*
*2.Department of Mechanical Engineering, The University of British Columbia, Vancouver V6T1Z4)*

**Abstract:** Environmental information can provide reliable prior information of human motion intent, which can aid the subject with wearable robotics to walk in complex environments. Previous researchers have utilized 1D signal and 2D images to classify environments, but they may face the problems of self-occlusion. Comparatively, 3D point cloud can be more appropriate to depict environments, thus we propose a directional PointNet to classify 3D point cloud directly. By utilizing the orientation information of the point cloud, the directional PointNet can classify daily terrains, including level ground, up stairs, and down stairs, and the classification accuracy achieves 99% for testing set. Moreover, the directional PointNet is more efficient than the previous PointNet because the T-net, which is utilized to estimate the transformation of the point cloud, is removed in this research and the length of the global feature is optimized. The experimental results demonstrate that the directional PointNet can classify the environments robustly and efficiently.

## 1 Introduction

Wearable robotics, including lower limb exoskeletons and prostheses, can assist millions of paraplegics and amputees in regaining walking ability[1]-[4]. Although humans can wear some wearable robotics to accomplish rhythmic locomotion, such as walking on the level ground or treadmill[5], they still face challenges to achieve non-rhythmic locomotion, like switching locomotion modes in complex environments[6]. The non-rhythmic locomotion is more difficult for the wearable robotics because most wearable robotics only focus on the kinetic or biological signals of human and lack the ability to perceive environments[7]-[11].

The able-bodied human can walk in complex environments based on a complete vision-locomotion loop. Human eyes can observe terrains in advance, and facilitate the brain to optimize the gait modes[12]. However, this vision-locomotion loop is incomplete for paraplegics and amputees. Consequently, researchers endeavor to add "eyes" to the wearable robotics and enhance the environmental adaptability of the wearable robotics. The depth camera, the laser sensor, the RGB camera, and the LIDAR have been utilized to recognize daily environments[13]-[17], including level ground, up stairs, and down stairs. Previous researchers have validated that the environmental recognition system can provide reliable prior information for the wearable robotics to switch locomotion modes and the environmental information is user-independent[18]. Nevertheless, there are still some limitations in previous research. The researchers have introduced many hyperparameters to extract reliable features from the origin environmental information. The environmental classification methods, like threshold method, decision tree, and support vectors machine, depend on the experience of the researchers and may not be robust nor general in real environments. Additionally, previous researchers only adopted 1D signal or 2D images to classify terrains, which may face the problem of self-occlusion. In consequence, an end-to-end environmental classification method based on 3D point cloud is more appropriate.

The 3D point cloud provided by the depth camera or LIDAR can depict daily environments well and can be applied to classify environments. However, there are some challenges to classify point cloud, including unordered and unstructured peculiarities. Traditionally, researchers inclined to map the unstructured point cloud to some structured spaces. For instance, the point cloud can be transformed to 3D voxel grids and be classified by 3D ShapeNets[19] and VoxelNet[20]. However, the voxelization methods may cause the curse of dimensionality and limit the resolution of the point cloud. For this reason, the voxelization methods are not satisfactory to process the point cloud of large size. Some researchers also attempted to project the point cloud into several planes and utilized multiple 2D CNN to recognize these 2D images, such as Multi-view CNN[21]. Nevertheless, the projection from 3D point cloud to 2D images may lose some critical information and face challenges to handle self-occlusion problems. Recently, PointNet, an end-to-end deep neural network, has been introduced to classify and segment the 3D point cloud directly[22]. The PointNet++ and frustum PointNets have also been proposed to extract local features and remove irrelevant regions by using corresponding RGB images[23], [24].

Because the global feature extracted by the PointNet is adequate to classify environments and the PointNet is more efficient than PointNet++ and frustum PointNet, we decide to utilize the PointNet to classify the daily environments (level ground, up stairs, and down stairs). Moreover, the invariance under transformations is inapplicable to the point cloud of environments. For instance, the transformations may mingle up stairs with down stairs. Under such circumstances, we apply a sensor fusion method[25] to combine an inertial measurement unit (IMU) with a depth camera to capture stable point cloud and simplify the PointNet to a directional PointNet.

In this research, we hypothesize that the directional PointNet can classify the stable 3D point cloud of daily environments (level ground, up stairs, and down stairs) directly, which is beneficial for the wearable robotics to achieve non-rhythmic locomotion in complex environments. The main contributions of our study include 1) introducing a directional PointNet to classify 3D point cloud of daily environments directly, 2) comparing the performance in this application between the directional PointNet and the PointNet, and 3) collecting the 3D point cloud dataset of the daily environments.

The rest of the paper is organized as follows. Section 2 describes the theoretical and experimental methods of our research. Experimental results are presented in section 3. Section 4 provides the corresponding discussions. The conclusions of the paper are presented in section 5.

## 2 Methods

The vision system and overall process of environmental classification are shown in Fig.1. The IMU and the depth camera are fixed on the belt, and they can provide a stable point cloud after synchronization. The original point cloud is dense but requires a large computational cost. Actually, there are superfluous points in the dense point cloud. Hence we downsampled the point cloud first. Then every point of the downsampled point cloud can be connected to the directional PointNet to extract features. The global features are extracted through multi-layer perceptron and are utilized to classify the point cloud directly.

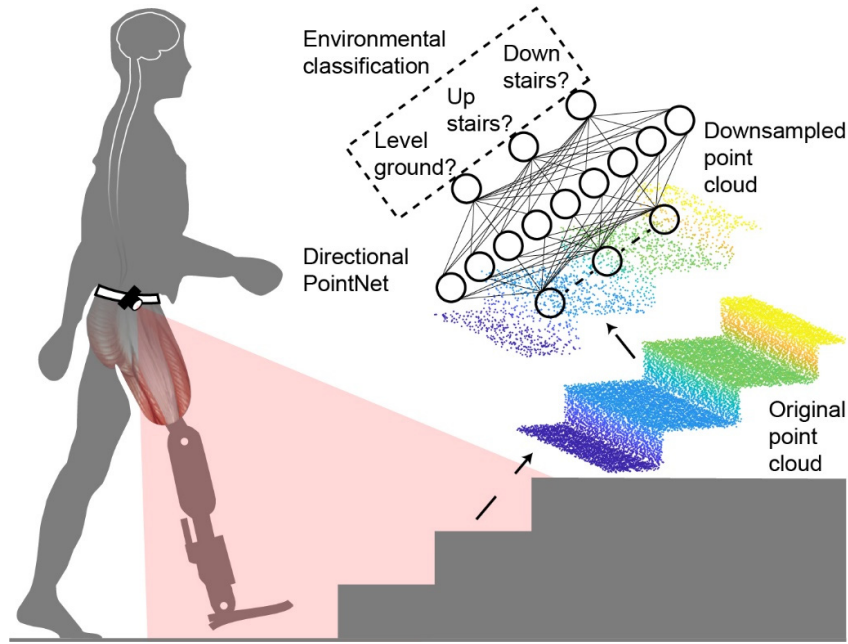

**Fig.1 The vision system and the overall process of environmental classification. The vision system is worn on the belt of the subject and provides the stable point cloud. The point cloud is downsampled first and is classified through a directional PointNet.**

## 2.1 Properties of Environmental Point Cloud

Similar to the point cloud used in the previous research[22], the environmental point cloud is also unordered. Hence the neural network should be invariant to the sequence variation of the input point cloud. However, the type of environmental point cloud depends on specific orientation, which is different from the invariance property of the point cloud in the previous research[22]. For instance, after rotation, the point cloud of up stairs can be transformed to the point cloud of down stairs. Consequently, the input point cloud should be stabilized first to distinguish between the point cloud of up stairs and down stairs. Moreover, the designed neural network should observe the rotation of the point cloud.

## 2.2 Point Cloud Stabilization

In our previous research, a point cloud stabilization method was presented[18]. Because the depth camera is worn on the belt, it will shake when a human is walking. In this case, the point cloud provided by the depth camera cannot be stable. In order to stabilize the point cloud, we used an IMU to measure the rotation angle of the camera and calculate the rotation matrix from the ground-based coordinate system to the camera-based coordinate system in real time. Then we transformed the point cloud to the ground-based coordinate system using the above rotation matrix.

## 2.3 Directional PointNet Architecture

The point cloud is unordered but direction-dependent. Thus we designed a directional PointNet (see Fig.2) based on the architecture of the PointNet[22].

In considering of the unordered property of the point cloud, we only select symmetrical functions: multi-layer perceptron and max pooling layer. The multi-layer perceptron (MLPs) connect with input points ($n \times 3$) and convert each point ($1 \times 3$) to a feature vector ($1 \times N$). Besides, the MLPs for different points share the parameters to ensure symmetry. The extracted feature matrix ($n \times N$) is aggregated to a global feature ($1 \times N$) through a max pooling layer. The combination of the shared MLPs and the max pooling layer is a symmetrical function and the variation of the sequence of input points will not influence the extracted global features. Finally, the global feature is utilized to calculate the classification scores for $k$ classes by another MLP.

Previous researchers designed a T-net to estimate the affine transformation matrix and offset the transformation of the point cloud[22]. However, this method is not appropriate for environmental classification because the rotation of the point cloud could affect the class of the point cloud, such as up stairs and down stairs. Therefore, we expunged the T-net and utilized an IMU to stabilize the point cloud instead.

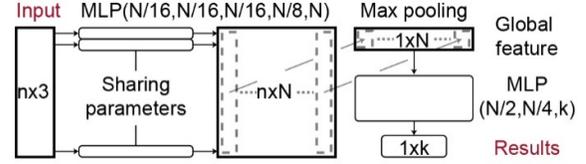

**Fig.2 Directional PointNet architecture. The input point cloud consists of $n$ points, and each point has three coordinates $x$, $y$ and $z$. The multi-layer perceptron (MLP) converts each point to a $1 \times N$ feature vector, and the MLPs for different points share parameters to ensure symmetry. The global feature is extracted through a max pooling layer and is utilized to calculate the classification results for $k$ classes by another MLP.**

## 2.4 Environmental Data Collection

The environmental dataset consists of simulated point cloud (50%) and the point cloud of the real environment (50%). The simulated point cloud was generated based on the general characteristics of the environments[18]. The point cloud of the real environment was captured by a depth camera, which is worn on the belt of a subject. As shown in Fig.3, there are three types of environments: level ground, up stairs, and down stairs.

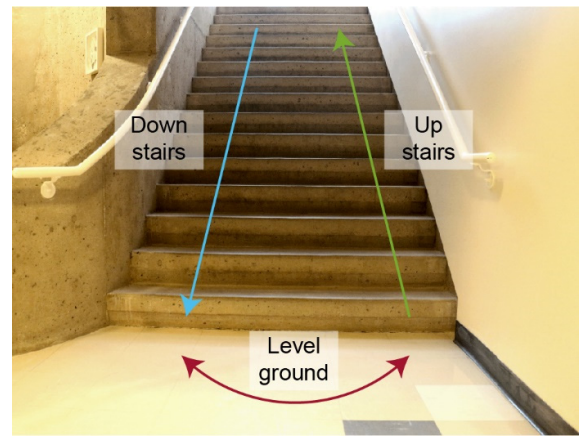

**Fig.3 Experimental environment.**

## 3 Results

There are 4016 point cloud samples from three ($k = 3$) different categories (level ground, up stairs, and down stairs) in our dataset. The overall dataset was split into a training set (50%) and testing set (50%). Each point cloud sample is composed of 2048 ( $n = 2048$ ) 3D points. During the training process, the initial learning rate, momentum, and batch size were set as 0.001, 0.9, and 32, respectively. The learning rate decayed during training, and the decay step and rate were 200000 and 0.7. The optimizer was selected as Adam optimizer.

### 3.1 Point Cloud of Different Environments

Three types of point clouds are shown in Fig.4 and the differences between different types of point cloud are discernible. Compared to the point cloud generated through the simulation, the point cloud of real environment is noisier. Moreover, there are some interferential points in the point cloud of real environment, including the side wall and human leg.

### 3.2 Results of Using Global Feature with Different Lengths

The environmental classification algorithm will be implemented on portable devices in real time to control the motion of wearable robotics in future. Hence the computational cost should be decreased. In previous research[22], researchers set the length of the global feature as 1024 to classify 40 types of models, which may be time consuming for portable devices. Fortunately, in this study, only three types of environments need to be classified and the feature-length could be decreased to reduce the computational cost.

In order to optimize the length of the global feature, we have trained the presented directional PointNet using the global features with different lengths, and the corresponding classification accuracies and loss values are shown in Fig.5. Except for the length of 32, the global feature with other lengths can achieve very high accuracy ( $> 99\%$ ) quickly, and the convergence speed increases with the increase of the length of the global feature. Because the convergence rates of the directional PointNet between using a global feature with the lengths of 256 and using longer global feature are similar, and the increasing of the length of the global feature will increase the computational cost, we chose 256 as the length of the global feature in this research.

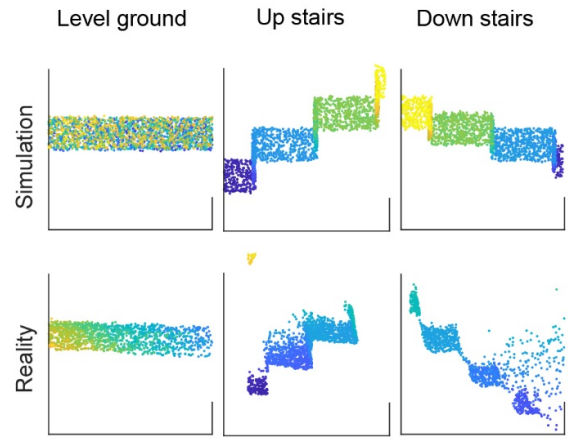

**Fig.4 Point clouds of different environments. The point clouds in the first row are generated through the simulation, and which in the second row are captured from the real environments**

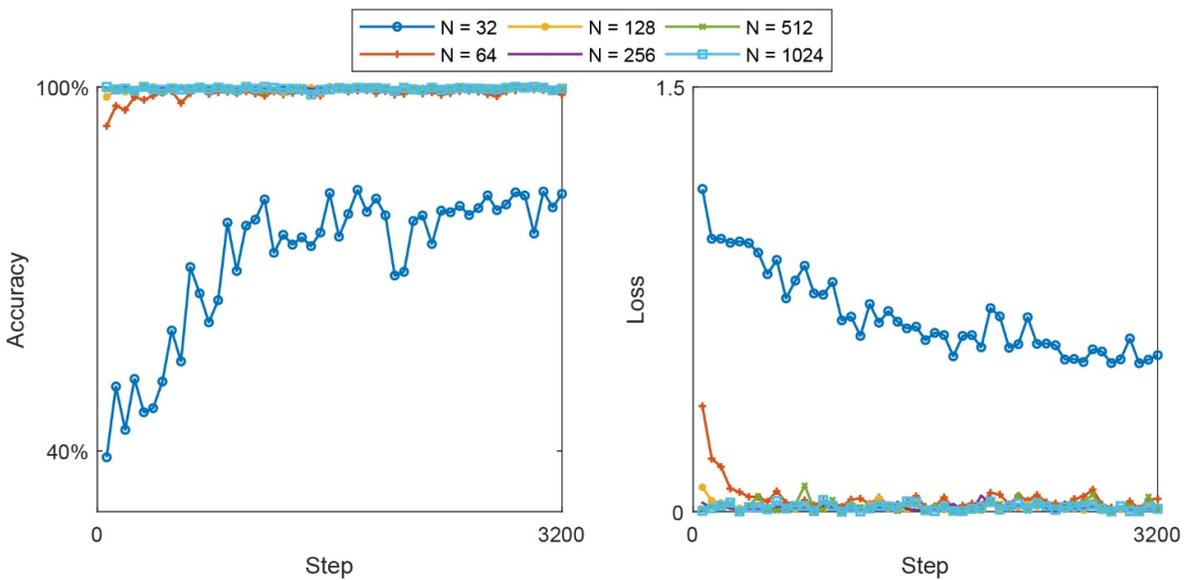

**Fig.5 Classification accuracy and loss values for testing set using the global feature with different lengths.** $N$ **is the length of global feature.**

### 3.3 Environmental Classification Results

The number of the global feature $N$ was set as 256 in this section, and the classification accuracy and loss values were adopted to evaluate the performance of the directional PointNet. Moreover, we also evaluated the performance of PointNet[22] on our dataset and compared the performance between our directional PointNet and the PointNet (see Fig.6). Compared to the PointNet, the classification accuracy of our directional PointNet increases more quickly, and the classification loss values of our directional PointNet decrease more quickly. The difference between out directional PointNet and the PointNet is that we remove the T-net. The results in Fig.6 validate that the T-net is not appropriate for the environmental classification in this research and it will decrease the convergence speed and increase computational cost.

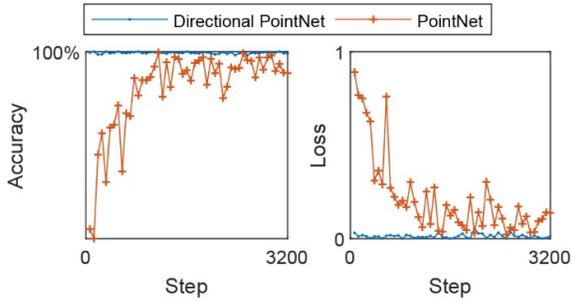

**Fig.6 Classification accuracy and loss values for the testing set.**

### 3.4 Visualizing Critical and Upper Bound Points

The global feature is just related to some points in a point cloud, which are called critical points. Moreover, the global feature will remain the same even after introducing some noisy points, and the points of the largest point cloud that has the same global feature as the critical points are call upper bound points.

Through calculating the corresponding points for the global feature, we can obtain the critical points. The upper bound points can be extracted from a $1m \times 1m \times 1m$ cube if the feature of the specific point is not larger than the global feature.

The results of critical and upper bound points are shown in Fig.7. Compared to original points and upper bound points, the critical points are sparse but can outline the important shape of different types of point cloud well. The upper bound points are dense and show the robustness of the presented method to deal with noise.

### 3.5 Computational Complexity Analysis

The training and testing of the directional PointNet were implemented on a computer with an Intel Core i7-6700 (3.4 GHz), a 16 GB DDR3, and a graphics card (GeForce GTX 1050 Ti). On this computer, the directional PointNet can classify the point cloud (2048 points/sample) quickly (2 ms/sample). We also compared the number of parameters and the floating-point operations/sample (FLOPs/sample) between the directional PointNet and the PointNet[22]. As shown in Table.1, the number of the parameters and the computational cost (FLOPs/sample) of the directional PointNet are much lower than the which of the PointNet because the T-net is removed in directional PointNet and the length of the global feature is decreased.

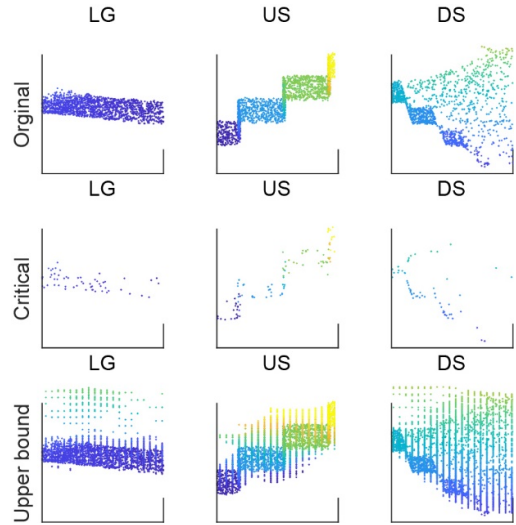

**Fig.7 Critical and upper bound points for different types of the point cloud. LG, US, and DS represent level ground, up stairs, and down stairs.**

Table.1 The number of parameters and computational cost (FLOPs/sample) for directional PointNet and PointNet.

| Methods | #Parameters | FLOPs/sample |
| --- | --- | --- |
| Directional PointNet | **0.05M** | **43M** |
| PointNet | 3.5M | 440M |

## 4 Discussion

In this research, we introduce a directional PointNet to classify daily terrains. For the directional PointNet, we remove the T-net because the orientation information is also important to classify different terrains, such as up stairs and down stairs. Moreover, we have combined an IMU with a depth camera to capture stable point cloud. Although the PointNet can also achieve high classification accuracy and low loss values, the computational cost of the PointNet is higher than the directional PointNet. Additionally, during training, the PointNet converges more slowly than the directional PointNet, because the PointNet needs to extract other deeper features except orientation information to classify up stairs and down stairs. Consequently, although the T-net can increase classification accuracy and promote the rotation invariance of the PointNet[22], it is futile to classify the daily terrains in this research.

We also increase the efficiency of the presented directional PointNet by optimizing the length of the global features. The experimental results show that the classification accuracy and loss values are improved infinitesimally after increasing the length of the global feature to larger than 256. The length of the global feature is relative to the number of the category of the point cloud. Previously[22], [23], researchers set the length of the global feature to 1024 to classify 40 types of the point cloud, because the critical global features for different categories are different and the neural network needs to extract all features simultaneously to classify all categories. Nevertheless, there are only three categories in this research, thus the feature-length of 256 is enough to classify these three types of daily terrains.

Furthermore, the critical and upper bound points are analyzed in this research. Only a few critical points can determine the category of a point cloud and the global feature will not be affected even adding some noisy points. Compare to our previous method[18], the directional PointNet could be more robust because it can classify 3D point cloud directly and can avoid self-occlusion problems. Moreover, the directional PointNet can also extract the principal components of the point cloud automatically, which can be utilized to estimate the environmental parameters.

The environmental classification accuracy is high in this study because the number of environmental categories is relatively small. Although these types environmental categories can also be classified by using traditional methods, including threshold, least-square, and support vector machine methods, the presented directional PointNet is more satisfactory because it avoids feature engineering and is more convenient to be applied to classify the point cloud of more categories.

Although the directional PointNet can classify the daily terrains with high accuracy, we acknowledge that there are still some limitations. Firstly, we should expand the categories of the point cloud, such as obstacle, ramp, and wall, to enhance the environmental adaptability of the wearable robotics in more complex environments. Besides, the environmental point cloud should also be segmented to estimate some parameters of the environments, which can be beneficial for the path planning of the wearable robotics. Finally, the method of the directional PointNet has only been evaluated in the offline analysis. We should apply this method to real-time control of the wearable robotics in the future to assess this method further.

# 5 Conclusion

In this study, we introduced a directional PointNet to classify 3D point cloud of daily terrains directly. The performance of the presented directional PointNet is evaluated through the offline classification experiments. The directional PointNet can classify different daily terrains accurately ( > 99%) and efficiently, and it can converge more quickly than the PointNet because it utilizes the orientation information of the point cloud rather than relies on a T-net. We also optimized of the length of the global features and visualized the critical and upper bound points to explain the outcome of our directional PointNet.

# References

[1] Young, A.J. and Ferris, D.P. (2017). State of the Art and Future Directions for Lower Limb Robotic Exoskeletons. *IEEE Transactions on Neural Systems and Rehabilitation Engineering*, 25(2), pp.171–182.

[2] Yan, T., Cempini, M., Oddo, C.M. and Vitiello, N. (2015). Review of assistive strategies in powered lower-limb orthoses and exoskeletons. *Robotics and Autonomous Systems*, 64, pp.120–136.

[3] Imam, B., Miller, W.C., Finlayson, H.C., Eng, J.J. and Jarus, T. (2017). Incidence of lower limb amputation in Canada. *Can J Public Health*, 108(4), pp.374–380.

[4] Ziegler-Graham, K., MacKenzie, E.J., Ephraim, P.L., Travison, T.G. and Brookmeyer, R. (2008). Estimating the Prevalence of Limb Loss in the United States: 2005 to 2050. *Archives of Physical Medicine and Rehabilitation*, 89(3), pp.422–429.

[5] Au, S.K., Weber, J. and Herr, H. (2009). Powered

Ankle–Foot Prosthesis Improves Walking Metabolic Economy. *IEEE Transactions on Robotics*, 25(1), pp.51–66.

[6] Varol, H.A., Sup, F. and Goldfarb, M. (2010). Multiclass Real-Time Intent Recognition of a Powered Lower Limb Prosthesis. *IEEE Transactions on Biomedical Engineering*, 57(3), pp.542–551.

[7] Bartlett, H.L. and Goldfarb, M. (2018). A Phase Variable Approach for IMU-Based Locomotion Activity Recognition. *IEEE Transactions on Biomedical Engineering*, 65(6), pp.1330–1338.

[8] Stolyarov, R., Burnett, G. and Herr, H. (2018). Translational Motion Tracking of Leg Joints for Enhanced Prediction of Walking Tasks. *IEEE Transactions on Biomedical Engineering*, 65(4), pp.763–769.

[9] Xu, D., Feng, Y., Mai, J. and Wang, Q. (2018). Real-Time On-Board Recognition of Continuous Locomotion Modes for Amputees With Robotic Transtibial Prostheses. *IEEE Transactions on Neural Systems and Rehabilitation Engineering*, 26(10), pp.2015–2025.

[10] Huang, S. and Huang, H. (2018). Voluntary Control of Residual Antagonistic Muscles in Transtibial Amputees: Feedforward Ballistic Contractions and Implications for Direct Neural Control of Powered Lower Limb Prostheses. *IEEE Transactions on Neural Systems and Rehabilitation Engineering*, 26(4), pp.894–903.

[11] Clites, T.R., Carty, M.J., Ullauri, J.B., Carney, M.E., Mooney, L.M., Duval, J.-F., Srinivasan, S.S. and Herr, H.M. (2018). Proprioception from a neurally controlled lower-extremity prosthesis. *Science Translational Medicine*, 10(443), p.eaap8373.

[12] Matthis, J.S., Yates, J.L. and Hayhoe, M.M. (2018). Gaze and the Control of Foot Placement When Walking in Natural Terrain. *Current Biology*, 28(8), pp.1224-1233.e5.

[13] Krausz, N.E., Lenzi, T. and Hargrove, L.J. (2015). Depth Sensing for Improved Control of Lower Limb Prostheses. *IEEE Transactions on Biomedical Engineering*, 62(11), pp.2576–2587.

[14] Liu, M., Wang, D. and Huang, H.H. (2016). Development of an Environment-Aware Locomotion Mode Recognition System for Powered Lower Limb Prostheses. *IEEE Transactions on Neural Systems and Rehabilitation Engineering*, 24(4), pp.434–443.

[15] Massalin, Y., Abdrakhmanova, M. and Varol, H.A. (2018). User-Independent Intent Recognition for Lower Limb Prostheses Using Depth Sensing. *IEEE Transactions on Biomedical Engineering*, 65(8), pp.1759–1770.

[16] Diaz, J.P., Silva, R.L. da, Zhong, B., Huang, H.H. and Lobaton, E. (2018). Visual Terrain Identification and Surface Inclination Estimation for Improving Human Locomotion with a Lower-Limb Prosthetic. In: *2018 40th Annual International Conference of the IEEE Engineering in Medicine and Biology Society (EMBC)*. 2018 40th Annual International Conference of the IEEE Engineering in Medicine and Biology Society (EMBC). pp.1817–1820.

[17] Pan, Y., Zhang, J., Song, L. and Fu, C. (2018). Real-time Terrain Mode Recognition Module for the Control of Lower-limb Prosthesis. In: *Proceedings of the 2018 International Conference on Computer Modeling, Simulation and Algorithm (CMSA 2018)*. 2018 International Conference on Computer Modeling, Simulation and Algorithm (CMSA 2018). Beijing, China: Atlantis Press.

[18] Zhang, K., Xiong, C., Zhang, W., Liu, H., Lai, D., Rong, Y. and Fu, C. (2019). Environmental Features Recognition for Lower Limb Prostheses Toward Predictive Walking. *IEEE Transactions on Neural Systems and Rehabilitation Engineering*, 2019, pp.1–11.

[19] Wu, Z., Song, S., Khosla, A., Yu, F., Zhang, L., Tang, X. and Xiao, J. (2015). 3D ShapeNets: A Deep Representation for Volumetric Shapes. In: Proceedings of the IEEE Conference on Computer Vision and Pattern Recognition. pp.1912–1920.

[20] Maturana, D. and Scherer, S. (2015). VoxNet: A 3D Convolutional Neural Network for real-time object

recognition. In: *2015 IEEE/RSJ International Conference on Intelligent Robots and Systems (IROS)*. 2015 IEEE/RSJ International Conference on Intelligent Robots and Systems (IROS). pp.922–928.

[21] Su, H., Maji, S., Kalogerakis, E. and Learned-Miller, E. (2015). Multi-view Convolutional Neural Networks for 3D Shape Recognition. In: *2015 IEEE International Conference on Computer Vision (ICCV)*. 2015 IEEE International Conference on Computer Vision (ICCV). pp.945–953.

[22] Qi, Charles R., Su, H., Kaichun, M. and Guibas, L.J. (2017). PointNet: Deep Learning on Point Sets for 3D Classification and Segmentation. In: *2017 IEEE Conference on Computer Vision and Pattern Recognition (CVPR)*. 2017 IEEE Conference on Computer Vision and Pattern Recognition (CVPR). pp.77–85.

[23] Qi, Charles R., Yi, L., Su, H. and Guibas, L.J. (2017). PointNet++: Deep Hierarchical Feature Learning on Point Sets in a Metric Space. In: Guyon, I. et al., eds. *Advances in Neural Information Processing Systems 30*. Curran Associates, Inc., pp.5099–5108.

[24] Qi, Charles R., Liu, W., Wu, C., Su, H. and Guibas, L.J. (2017). Frustum PointNets for 3D Object Detection from RGB-D Data. *arXiv:1711.08488 [cs]* [online], 22 November 2017.

[25] Silva, C.W. de. (2017). Sensor Systems: Fundamentals and Applications. *Taylor & Francis/CRC Press*.

## Author Biographies

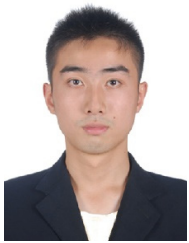

**KUANGEN Zhang** received the B.E. degree from Tsinghua University in 2016. Now he is a joint Ph.D. student of the University of British Columbia (UBC) and Southern University of Science and Technology (SUSTech). His research interests include robotic vision, sensor fusion, and wearable robots.

Email：kuangen.zhang@alumni.ubc.ca

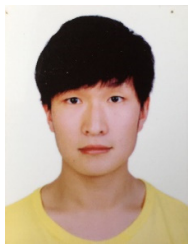

**JING Wang** received the B.ASc degree from The University of British Columbia (UBC) in 2018. Now he is an M.ASc student at the University of British Columbia. His research interests include computer vision and intelligent robotics.

Email:j.wang94@alumni.ubc.ca

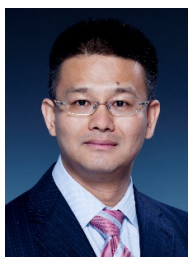

**CHENGLONG Fu** received the B.S. degree from the Department of Mechanical Engineering, Tongji University, Shanghai, China, and the Ph.D. degree from the Department of Precision Instruments and Mechanology, Tsinghua University, Beijing, China, in 2002 and 2007, respectively. From 2007 to 2010, he was an Assistant Professor with the Department of Precision Instruments and Mechanology, Tsinghua University. From 2011 to 2012, he was a Visiting Scholar with the Department of Mechanical Engineering, University of Michigan, Ann Arbor, USA. From 2010 to 2017, he was an Associate Professor with the Department of Mechanical Engineering, Tsinghua University. He is currently an Associate Professor with the Department of Mechanical and Energy Engineering, Southern University of Science and Technology, Shenzhen, China. His research interests include wearable robots and human-centered robotics.

Email：fucl@sustc.edu.cn